# Anti-Jamming Path Planning Using GCN for Multi-UAV

Haechan Jeong [1]

[1]The CCS Graduate School of Mobility, Korea Advanced Institute of Science and Technology (KAIST), Daejeon 34051, Republic of Korea

*Abstract—* **This paper addresses the increasing significance of UAVs (Unmanned Aerial Vehicles) and the emergence of UAV swarms for collaborative operations in various domains. However, the effectiveness of UAV swarms can be severely compromised by jamming technology, necessitating robust anti-jamming strategies. While existing methods such as frequency hopping and physical path planning have been explored, there remains a gap in research on path planning for UAV swarms when the jammer's location is unknown. To address this, a novel approach, where UAV swarms leverage collective intelligence to predict jamming areas, evade them, and efficiently reach target destinations, is proposed. This approach utilizes Graph Convolutional Networks (GCN) to predict the location and intensity of jamming areas based on information gathered from each UAV. A multi-agent control algorithm is then employed to disperse the UAV swarm, avoid jamming, and regroup upon reaching the target. Through simulations, the effectiveness of the proposed method is demonstrated, showcasing accurate prediction of jamming areas and successful evasion through obstacle avoidance algorithms, ultimately achieving the mission objective. Proposed method offers robustness, scalability, and computational efficiency, making it applicable across various scenarios where UAV swarms operate in potentially hostile environments.**

## I. INTRODUCTION

Recently, the importance of UAVs (Unmanned Aerial Vehicles) has been increasing. UAVs are being used in various fields such as natural disaster investigation [1], environmental monitoring [2], crop monitoring in agriculture[3], and infrastructure inspection[4]. They are also gaining significant importance in the defense sector, where they are used for various missions such as reconnaissance, strikes, and communication support in modern warfare [5]. Through these applications, UAVs can perform tasks that were previously difficult and enable efficient data collection and operation.

To maximize the potential of UAV applications, UAV swarms have emerged, where multiple UAVs collaborate and operate together as a single cluster to maximize efficiency [6]. Flocking is one of the important concepts in controlling UAV swarms. It requires multiple wireless connections [7] among UAVs and simple data transaction [8]. Flocking models the flocking behavior observed in nature, where multiple UAVs maintain connectivity and form a cohesive formation or alignment through interaction [9]. Therefore, flocking is considered a crucial concept for the efficient operation of UAV swarms.

However, the missions conducted by UAV swarms can be severely impacted by jamming technology [10]. Jamming is a technique that disrupts communication and makes information exchange and mission execution impossible for UAVs. In the case of UAV swarms, where multiple UAVs exchange information and perform collective missions, jamming becomes an even more critical issue. Therefore, anti-jamming technology is essential in UAV swarms to mitigate the effects of jamming.

Anti-jamming techniques for UAVs have been extensively studied and discussed in numerous research papers. One commonly used method is frequency hopping, where UAVs switch frequencies to avoid jamming [11]. However, processing input signals at every moment can be inefficient for UAVs due to their limited battery capacity. As an alternative, the physical movement of UAVs can be utilized. Particularly, previous studies have explored the use of physical path planning to avoid jamming areas, as the strength of jamming signals exponentially decreases with distance [12]. However, some of these studies assume that UAVs have knowledge of the jammer's location, which may not be realistic in practical situations. UAVs often do not have information about the jammer's location. Therefore, there is limited research on path planning for UAV swarms to evade jammers when the jammer's location is unknown.

In summary, UAVs have gained importance in various fields, and UAV swarms have emerged as a way to maximize their potential through collaborative operations. Flocking is a crucial concept for controlling UAV swarms effectively. However, the impact of jamming technology poses a significant threat to UAV swarms, making anti-jamming technology essential. Previous research has explored anti-jamming techniques for UAVs, including frequency hopping and physical path planning. However, there is a need for further research on path planning for UAV swarms to evade jammers when the jammer's location is unknown.

This paper introduces a novel approach where UAV swarms utilize collective intelligence to predict jamming areas, evade them, and quickly reach target destinations to reestablish flocking. Based on mathematical modeling of jamming signals, a simulation model is constructed to generate information about the jammer's location and the intensity of jamming signals received by each UAV at their current positions. The proposed method utilizes Graph Convolutional Networks (GCN) to gather information from each UAV, predict the location and intensity of the jamming area at each time step. Using the predicted information about the jamming area, a multi-agent control algorithm is employed to disperse the UAV swarm, avoid jamming, and regroup upon reaching the target destination.

This approach contributes in the following ways. Firstly, even in situations where the information about the jammer's location and intensity is unknown in advance, this method can predict them based on the given information, making it robust and widely applicable. Secondly, by leveraging the algorithm structure of GCN and utilizing collective intelligence from the information each UAV possesses, this framework achieves accuracy in prediction. Thirdly, by utilizing the physical movement of UAVs to evade jamming, effective anti-jamming is achieved. Lastly, by using multi-agent control, it enables the simultaneous control of multiple UAVs, which reduces computational costs and allows for fast prediction of jammers.

The rest of the paper is structured as follows. Section II describes the problem scenario and jamming modeling. Section III introduces the multi-agent algorithm and GCN network. Section IV explains the simulation setup and presents the results. Finally, Section V provides the conclusion.

## II. PROBLEM STATEMENT

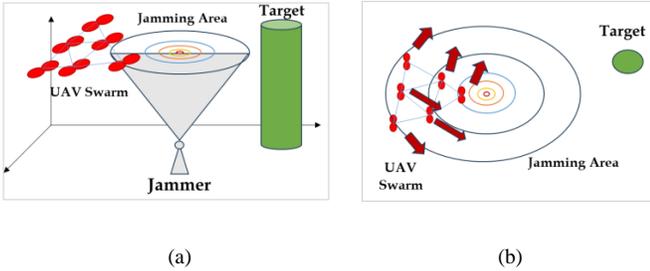

Fig. 1. Overview of the simulation environment: (a) situation where a multi-UAV consisting of six UAVs is flying in a triangular formation towards the target;(b) situation as Fig. 1(a) but viewed from the ground. The jamming area represents the region affected by jamming signals emitted by the jammer. Within the jamming area, the strength of the jamming signal significantly affects the communication status of UAVs, with greater impact observed closer to the center.

While flying as shown in Figure 1, UAVs encounter a jamming area formed by jammers. To avoid jamming and reach the target, each UAV disperses according to the path planning derived from their respective routes. After avoiding the jamming area, the dispersed UAVs gather at the target to reform the formation, which is their common objective. It can be observed that the probability $P$ of UAV experiencing communication disruption due to the jammer is inversely related to the distance $r$ between the center of the jamming area and UAV. For the sake of simplicity, this relationship can be modeled using the following equation:

$$P = kA^r, \quad (0 < P < 1, 0 < A < 1) \qquad (1)$$

where $k$ and $A$ are constants that vary depending on the signal strength of the jammer. When $P$ exceeds a certain level, the communication of the UAV is completely disrupted, and it remains within the jamming area without being able to move. This threshold value is denoted as $P_\tau$, and the distance $r_\tau$ between the jammer and the UAV when $P$ reaches this value is given by

$$r_\tau = \log_A \frac{P_\tau}{k} \qquad (2)$$

Therefore, UAVs must satisfy $r > r_\tau$ during flight in order to avoid experience severe communication disruptions.

## III. PROPOSED METHOD

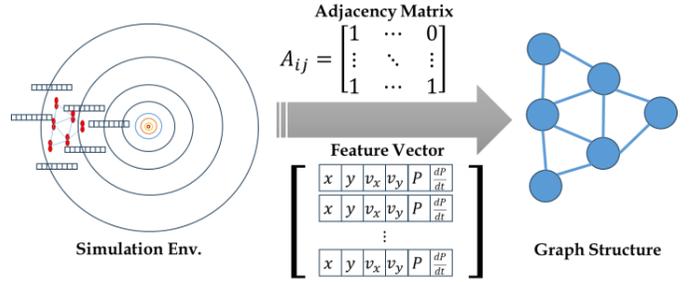

Fig. 2. Simulation scenario represented as a graph structure where each node corresponds to a UAV. The indices $i$ and $j$ represent the indices of the UAVs. The adjacency matrix is a binary matrix where each element indicates whether the $i$-th UAV and the $j$-th UAV are within the maximum communication range (1 if yes, 0 otherwise). The feature vector consists of the current position, velocity, probability $P$, and the time derivative of $P$ for each UAV.

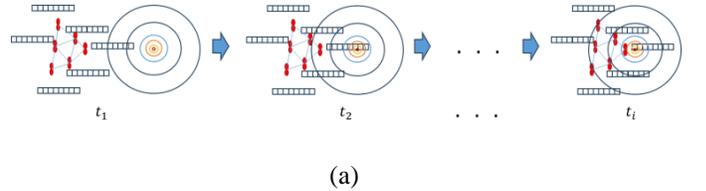

(a)

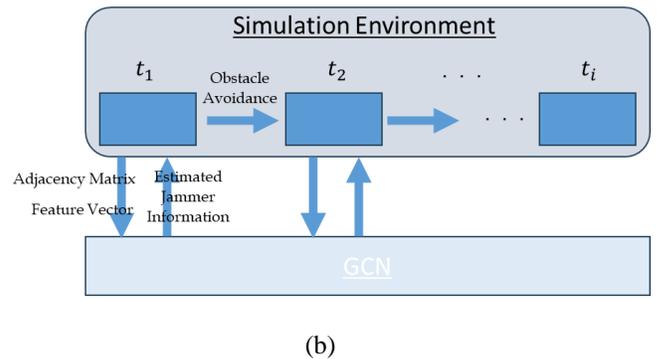

(b)

Fig. 3. Interaction between the UAV swarm and the jamming area and simulation environment:(a) interaction between the UAV swarm and the jamming area;(b) overview of the algorithm's structure. The situation of the UAV swarm is transformed into a graph structure and fed into the GCN as input. The trained GCN provides predictions of the jammer's information, including the jammer's location and the intensity of the jamming signal, based on this input. If the predicted values indicate that a danger zone is approaching the UAVs, the UAVs regenerate routes to circumvent this area using the obstacle avoidance algorithm. This process is repeated at regular time intervals until the UAV swarm reaches its destination.

The GCN is a specialized network for processing data with graph structures. As each layer of GCN is traversed, it updates the feature vectors of each node by incorporating information from neighboring nodes through interactions, gradually reflecting the global information of the input data. In other words, GCN layers provide the advantage of effectively learning patterns in graph-structured data by utilizing neighbor aggregation. It is possible to match each UAV with a node in the graph to form a graph structure. Use of GCN can be effective for the framework of reinforcement learning [13]. In other words, in the simulation, the $i$-th UAV becomes the $i$-th node in the graph. The graph consists of six nodes, e.g., six UAVs. Two UAVs that are closer than the maximum communication range $d$ are considered to be connected. Based on this, the distance between each pair of UAVs can be calculated to obtain the adjacency matrix $A_{ij}$. By using equation (1), the $i$-th UAV can calculate its own jammer-induced power $P_i(t)$ and $\frac{dP_i(t)}{dt}$ at time $t$ based on its position $(x_i(t), y_i(t))$ and velocity $(v_{xi}(t), v_{yi}(t))$. Therefore, a feature vector consisting of $x_i(t), y_i(t), v_{xi}(t), v_{yi}(t), P_i(t), \frac{dP_i(t)}{dt}$ can be obtained. As shown in the figure, using the acquired $A_{ij}$ and feature vectors from the simulation environment as input data, the GCN learns to accurately predict the jammer's position $(x_j, y_j)$ and the values of constant $A$ related to the jamming signal strength. Once trained, the GCN can predict $x_j, y_j, A$ from $A_{ij}$ and the feature vector. As shown in the figure, since the UAVs move at each moment, $A_{ij}$ and the feature vector are recalculated as inputs to GCN, and GCN predicts information about the jammer at each iteration. Based on the predicted information, the multi-UAV system avoids the region defined by $r_\tau$ through obstacle avoidance algorithms.

IV. EXPERIMENTS

*A. Setup*

The parameters modeling the jammer are set as follows: the threshold value, mentioned in Section II, $P_\tau$ is set to 0.5, and for simplicity, $k = 1$ is fixed. The parameters related to the multi-UAV dynamic systems are set as $d = 20$ m and $r = 1.2 \times d = 24$ m. Based on this setup, a simulation environment was constructed. From this environment, a combination of ground truth data for $x_j, y_j, A$, and the input data $A_{ij}$ and feature vectors were generated, totaling 100,000 samples. This dataset was used to train GCN. The hyperparameters related to GCN were set as follows: Batch size = 32, learning rate = 0.001, and the number of graph convolution layers = 2.

*B. Results*

Figure 4 depicts the training and validation loss obtained while training GCN for 1000 epochs using the dataset. It can be observed that both losses steadily decrease and converge as the training progresses. Figure 5 depicts the simulation results of the UAV swarm forming a cluster and navigating around the jammer towards the target location. Figures 5(a-h) show snapshots of the positions of the UAVs saved every 10 seconds.

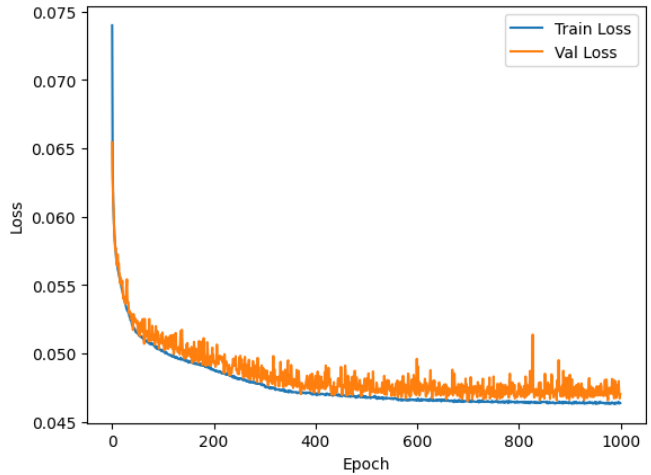

Fig. 4. Training and validation loss over 1000 epochs.

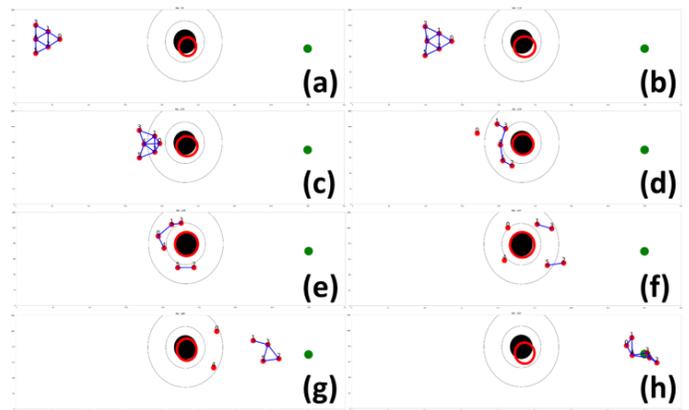

Fig. 5. Circular black lines in each Fig. 5(a)-(h) represent the contours of $P$ affected by the jamming signal. As the UAVs approach the centers of these concentric circles, the value of $P$ increases. The circles filled with black color represent areas where the probability $P$ exceeds $P_\tau$ due to the jamming signal, with the radius of the circle denoting $r_\tau$. UAVs must navigate around this area to maintain smooth communication. The red circles represent the predicted position of the jammer and the corresponding jamming signal strength as estimated by the multi-UAV system. The red dots represent the UAVs. The blue lines represent communication links, indicating that the two UAVs connected by this line are positioned closer to each other than the maximum communication range. The green circles represent the target location.

The predicted positions of the jammer by the trained GCN are represented as red circles in Fig. 5(a)-(h). In Fig. 5(a) and Fig. 5(b), the predictions are relatively less accurate because the jammer and the UAVs are far apart, resulting in weaker effects of the jamming signal. The elements of the feature vector, $P$ and $\frac{dP}{dt}$, are too small, leading to lower prediction accuracy. However, in such situations, there is no necessity for the UAV swarm to precisely locate and avoid the jammer, so an approximate estimation of the jammer's position suffices. As seen in Figs. 5(c-f), as the UAV swarm approaches the jammer, the predictions become more accurate. During this period, due to significant changes in the magnitude and rate of change of the jamming signal, the accuracy of GCN predictions regarding the jammer presence improves with the UAV swarm's position changes. Particularly in Fig. 5(d) and

(e), the UAVs that almost accurately predict the jammer's position disperse to avoid areas where $P$ exceeds $P_\tau$ for communication quality, implementing such movements with the obstacle avoidance algorithm within each UAV's path planning. In Fig. 5(g-h), UAVs safely move away from the danger zone, regroup within the range of mutual communication, and successfully reach the target location to complete the mission. Although the UAV swarm's prediction accuracy regarding the jammer decreases as they move further away from the jammer, it does not affect the overall mission significantly because of the mission already performed safely,.

## V. CONCLUSION

This paper demonstrates the use of GCN based on UAV information to accurately predict the jamming area with insufficient information and describes anti-jamming path planning using obstacle avoidance algorithm to reach the target. All scenarios are implemented through simulations based on modeling. Data are collected based on these simulations, and it is observed that the loss decreased and converged as the GCN is trained. Using the trained network, simulations are conducted to demonstrate that GCN accurately predicts the jammer's information and successfully avoids the jamming area through obstacle avoidance algorithms, ultimately reaching the target. This approach actively leverages the movement capabilities of UAVs to propose algorithms for avoiding jamming. Additionally, GCN enables the aggregation of information from multiple UAVs, allowing the utilization of collective knowledge and achieving high accuracy.